\documentclass{article}
\usepackage[T1]{fontenc}
\usepackage{hyperref}
\usepackage[utf8]{inputenc}
\usepackage{aux/spconf}
\usepackage{amsmath,graphicx}
\usepackage{lipsum}
\usepackage{caption}
\usepackage{threeparttable}
\usepackage{tabularx}
\usepackage{booktabs}
\usepackage[noadjust]{cite}
 
\usepackage{filecontents}

\title{Leveraging SLIC Superpixel Segmentation and Cascaded Ensemble \\ SVM for Fully Automated Mass Detection in Mammograms}

\name{Jaime Simarro, Zohaib Salahuddin, Ahmed Gouda, Anindya Shaha}
\address{
Università degli Studi di Cassino e del Lazio Meridionale, Italy \\
}

\begin{document}
\maketitle

\begin{abstract}
Identification and segmentation of breast masses in mammograms face complex challenges, owing to the highly variable nature of malignant densities with regards to their shape, contours, texture and orientation. Additionally, classifiers typically suffer from high class imbalance in region candidates, where normal tissue regions vastly outnumber malignant masses. This paper proposes a rigorous segmentation method, supported by morphological enhancement using grayscale linear filters. A novel cascaded ensemble of support vector machines (SVM) is used to effectively tackle the class imbalance and provide significant predictions. For True Positive Rate (TPR) of $0.35$, $0.69$ and $0.82$, the system generates only $0.1$, $0.5$ and $1.0$ False Positives/Image (FPI), respectively.

\end{abstract}

\begin{keywords}
mammogram, mass detection, superpixel, grayscale filter, support vector machine, cascade, ensemble
\end{keywords}

\vspace{6mm}

\section{Introduction}
Breast cancer represents the most frequent malignancy among women with an estimated $2.1$ million new cases and over $600,000$ deaths worldwide in 2018 \cite{1}. Early detection is critical and can ensure $5$-year survival rates of $85-99\%$, however only $62\%$ of cases are diagnosed at this stage \cite{2}. A common precursor to invasion is the manifestation of dense, high-intensity lumps with variable size and texture, termed "masses" \cite{3}. Using advanced image analysis and supervised machine learning algorithms, masses can be swiftly and effectively detected with a significant degree of accuracy in screening mammography samples \cite{4,5}.

In this paper, we propose an end-to-end integrated system to identify, segment and classify candidate masses in a mammogram by utilizing their fundamental features with regards to scale, topology, texture and relative local intensity as priori in every stage of the pipeline. Complete training and evaluation of the model is performed using the publicly available \textit{INbreast} \cite{6} full-field digital mammographic dataset, containing $115$ cases and $410$ instance-level annotated images at $70$ $\mu$m resolution.

\vfill

\begin{figure*}[t!]
\centerline{\includegraphics[width=\textwidth]{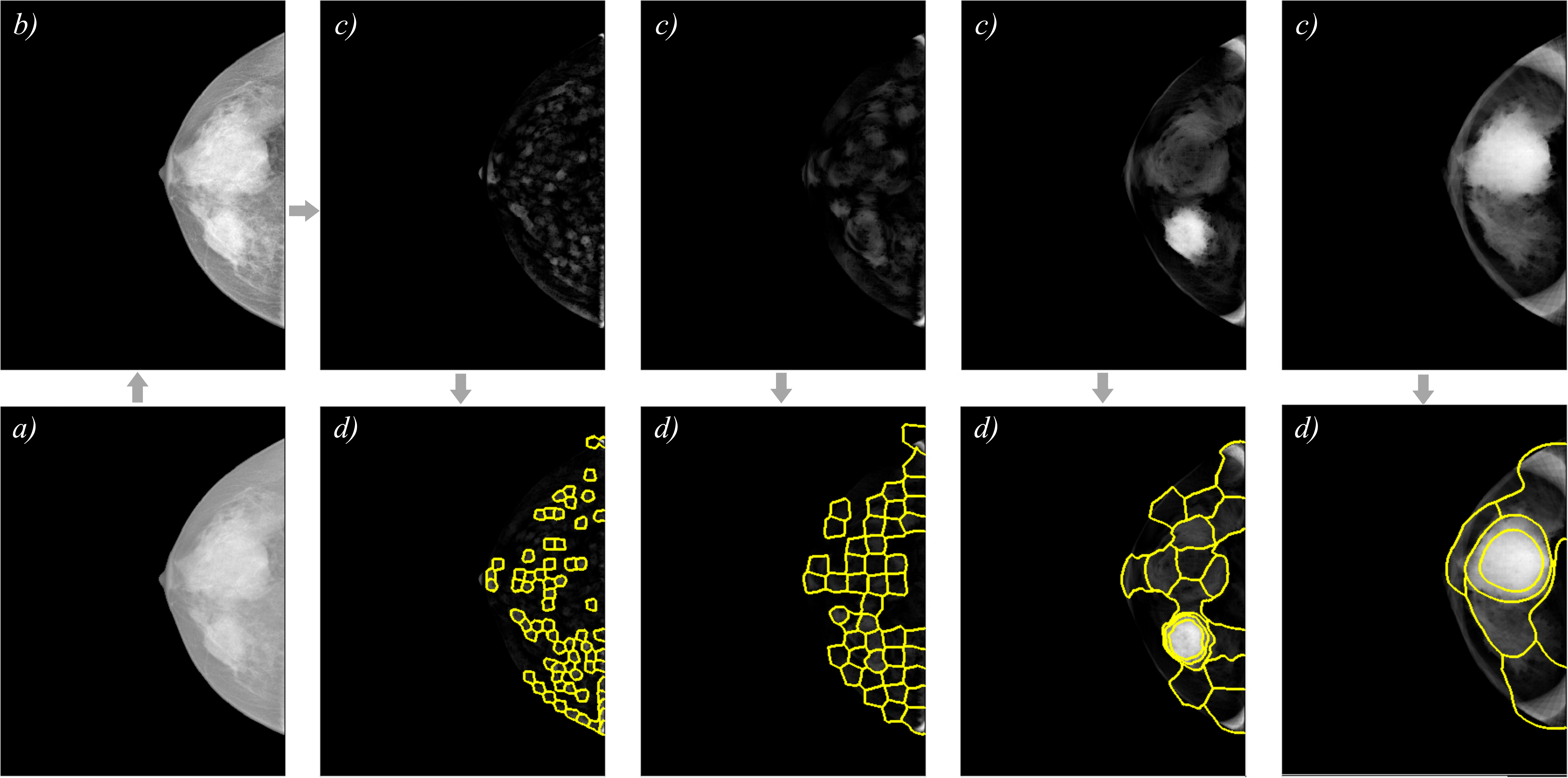}}
\caption{Region segmentation pipeline to identify candidate masses of variable size and shape: $a)$ \textit{Original Image}: $16$-bit screening mammogram resized by a scale factor of $4$; $b)$ \textit{Pre-Processed Image}: CLAHE with clipping limit $1.00$ and tile size $[4,4]$; $c)$ \textit{Enhanced Images}: Multi-scale morphological sifting for $18$ angular orientations per structuring element in $4$ scales; $d)$ \textit{Clustered Images}: SLIC superpixel segmentation with adaptive compact factor for each scale and smoothing $\sigma=5.00$.}
\end{figure*}

\section{Methodology}
\subsection{Region Segmentation}
\subsubsection{Pre-Processing and Equalization}
Digital mammography is primarily susceptible to heterogeneous, intensity-dependent quantum noise that limits its high-resolution expression. As such, an equalization or denoising technique is usually required to compensate for the same \cite{7}. We extract the breast using its corresponding mask (as provided in the dataset), linearly rescale its contrast range to $16$-bit depth, downsize the resultant image by a scale factor of 4 and apply \textit{Contrast Limited Histogram Equalization (CLAHE)}, as shown in Fig. 1.

\subsubsection{Multi-Scale Morphological Sifting}
Malignant masses can have irregular shapes with ambiguous boundaries. Smaller masses are nearly indistinguishable, if embedded in parenchymal tissue or in the absence of a central, consolidated density. However, they are often surrounded by stellate patterns of linear spicules \cite{9}. \textit{INbreast} contains $116$ masses, each covering a total cross-sectional area between $15-3689$ mm$^2$ \cite{4,6}. More generally, mass sizes can vary in the range $3-30$ mm in diameter \cite{8}, which translate to approximately $43-429$ pixels at $70$ $\mu$m resolution.

By exploiting these features, a pair of rotating line segments mapped in multiple orientations, at multiple scales (each limited to an appropriate range of diameters), can be used to design grayscale morphological filters. Dual top-hat operation using these structuring elements can simultaneously extract patterns of interest (corresponding to mass densities of different scales) and suppress background tissue \cite{4} in the image. Fig. 1 demonstrates how multiple masses of different sizes, in the same mammogram, can be effectively enhanced using this multi-scale approach.

\subsubsection{SLIC Superpixel Segmentation}
Superpixels can cluster perceptually meaningful information as opposed to a discretized arbitrary pixel-grid. Hence, we apply a \textit{Single Linear Iterative Clustering} (SLIC) algorithm, based on $k$-means, to group enhanced objects in each sifted image \cite{5} into superpixels sharing correlated intensity and spatial locality. Each superpixel is overlaid with the ground truth and labelled as a positive/negative candidate based on minimum degrees of intersection (evaluated using Dice coefficient) specified at each scale. When calibrated correctly, all masses are segmented as positive candidates with a moderate Dice coefficient ($>45\%$), but a large number of false candidates are generated as well. We apply a threshold to remove clearly redundant superpixels that are either outside the breast or too dark relative to the intensity profile of its given scale. This step alleviates the heavy class imbalance faced in the classification stage by an average factor of $10$ per image.

\vfill

\subsection{Mass Classification}
\subsubsection{Feature Extraction}
$23$ total unique features (as listed in Table 1) are computed for each superpixel incoming from the previous stage to discriminate between false and true candidates for masses. While some are fundamental, pertaining to shape, contours, texture and orientation, others are derived from base features and provide useful information to separate the two classes. Furthermore, certain features are computed for superpixels extracted from the CLAHE processed images and others are also computed for superpixels extracted from the multi-scale morphologically enhanced images.

\begin{table}[!h]
\renewcommand{\arraystretch}{1.3}
\centering
\caption{Discriminating Features in Classification}
\label{tab1}
\begin{threeparttable}
\begin{tabular}{lcr}
\toprule
\midrule
Skew  &            Mean  &                     Angular Moment \\
Circularity &      \hspace{0.6cm}              Smoothness \hspace{0.6cm} &     GLCM \\
Perimeter   &      Uniformity  &               Dissimilarity \\
Kurtosis  &        Entropy  &                  Homogeneity \\
Solidity  &        Contrast  &                 $(S_1-S_2)/S_1$\textsuperscript{1}    \\
Eccentricity  &    Radius  &                   $(S_2-S_3)/S_2$\textsuperscript{1}    \\
Extent  &          Area  &                     Isocontours\textsuperscript{2}    \\
Energy  &          Correlation  &              \\
\bottomrule
\end{tabular}
\begin{tablenotes}
   \item[1] \textit{Chu, J., Min, H. et al.} [5]
   \item[2] \textit{B.W. Hong, B.S. Sohn} [10] (conditional feature)
\end{tablenotes}
\end{threeparttable}
\end{table}

\subsubsection{Cascaded Ensemble of Support Vector Machines}
For the classifier, we propose a cascaded ensemble of support vector machines (SVM) in three stages. Its schematic architecture can be seen in Fig. 2. For the first and second stages of the cascade, we divide the negative candidates feature vector ($F$) into equal partitions and feed a single, unique partition to each SVM, paired with the entire positive candidates feature vector ($T$). In other words, the class imbalance is distributed across the ensemble of $N$ SVMs. Here, $N$ represents the ratio of $F:T$ divided by $10$. From the aggregated output at each stage, we use the true positives (TP) and false positives (FP) to construct the next $T$ and $F$, respectively, while discarding the true negatives and false negatives. This step further dilutes the imbalance down the cascade. The final stage comprises of a single SVM and its output (TP, FP, TN, FN) accounts for our predicted results.

\begin{figure}[t!]
\centerline{\includegraphics[width=0.35\textwidth]{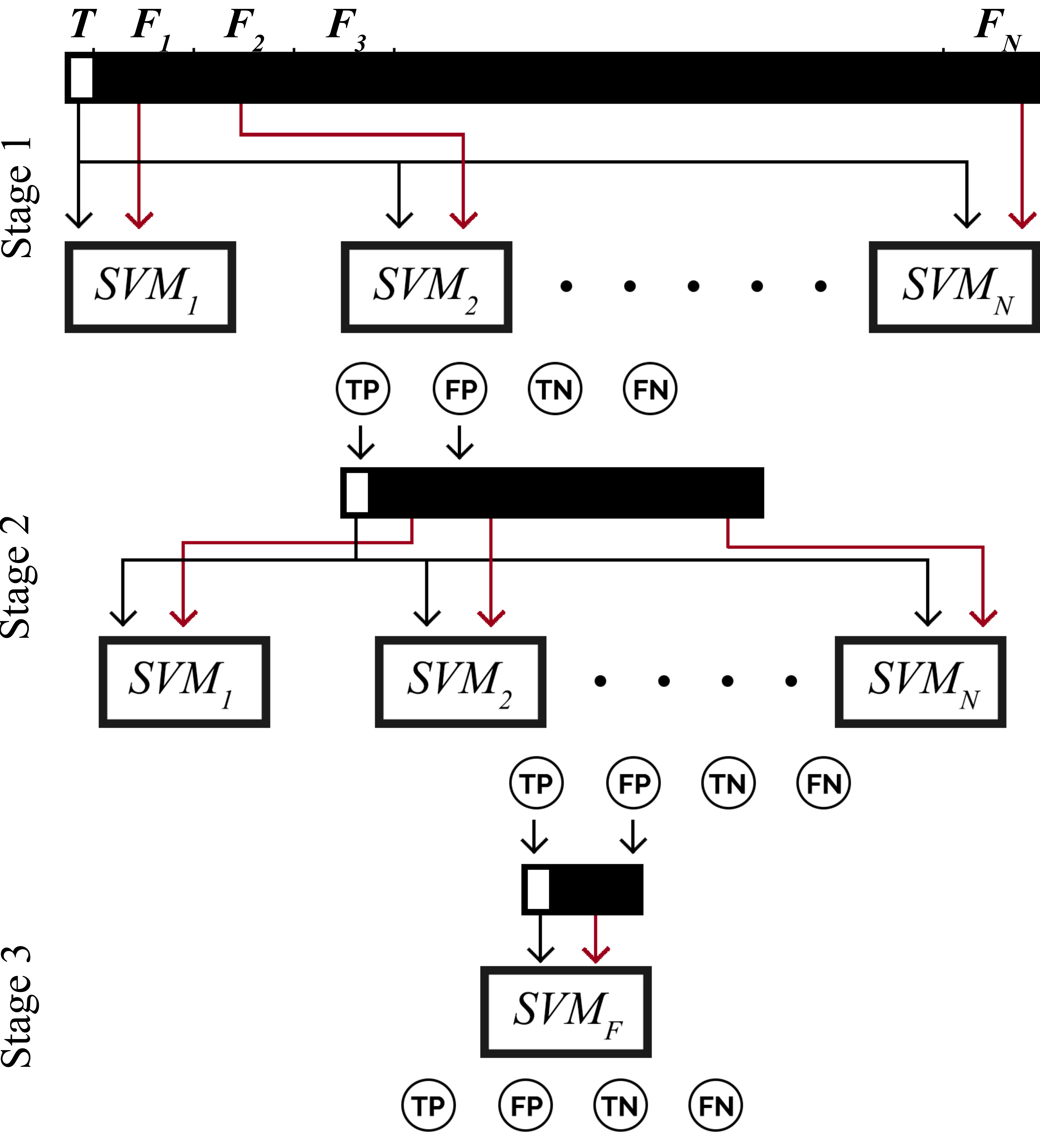}}
\caption{Architecture for cascaded ensemble of SVMs.}
\end{figure}

\vspace{4mm}

\section{Experimental Results}
The model is evaluated via $10$-fold cross-validation on the entire dataset of $410$ images, using a validation split of $10\%$. A minimum Dice coefficient of $0.20$ is required to verify a positive prediction as a true positive.

\begin{figure}[h!]
\centerline{\includegraphics[width=0.40\textwidth]{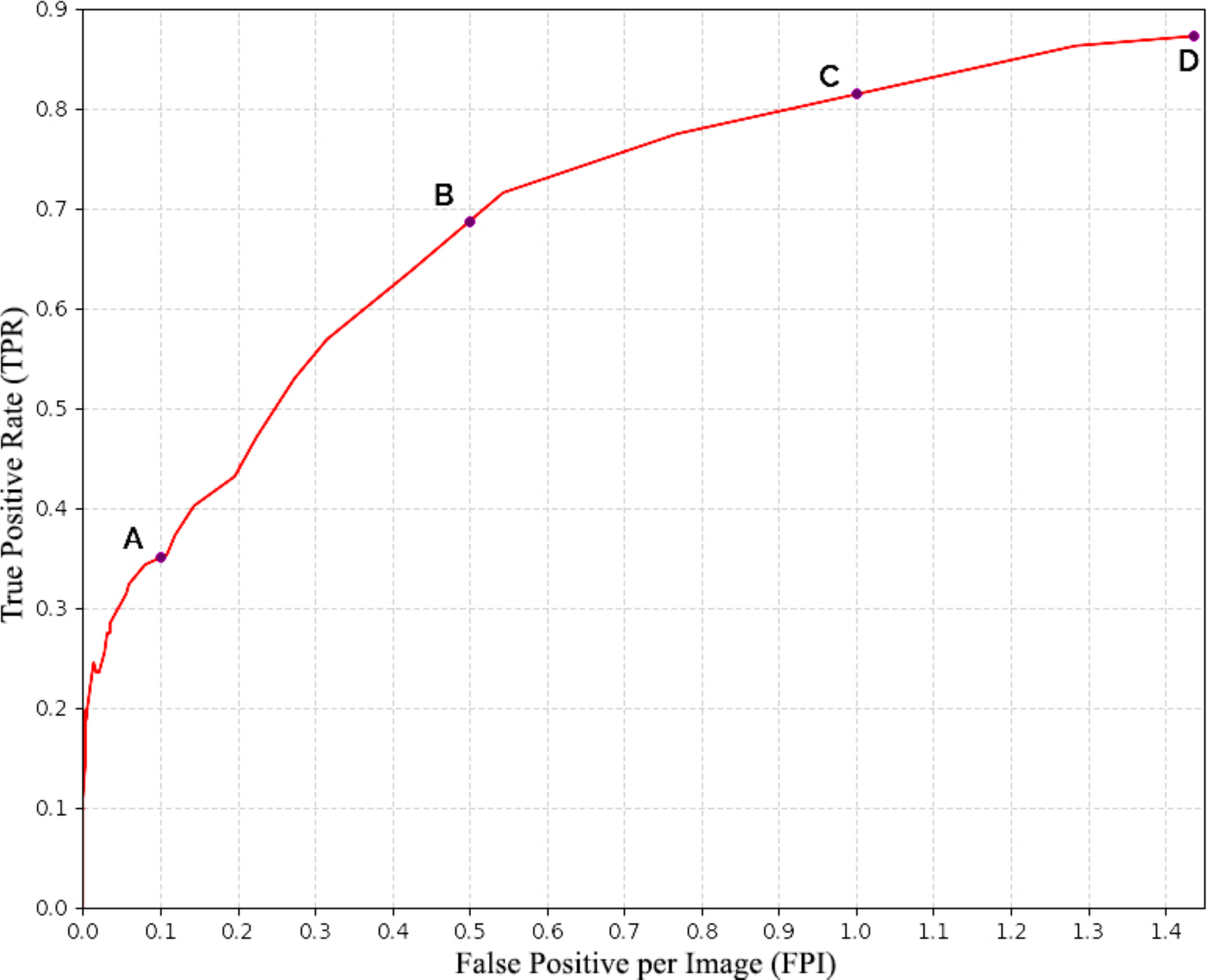}}
\caption{Free-response Receiver Operating Characteristic (FROC) curve: $A$, $B$, $C$ represent useful operating points of interest for radiologists, yielding TPR $0.35$, $0.69$, $0.82$ at FPI $0.1$, $0.5$, $1.0$, respectively. Point $D$ represents notable sensitivity (TPR $0.87$) within a reasonable specificity (FPI $1.44$).}
\end{figure}

\begin{figure}[h!]
\centerline{\includegraphics[width=0.48\textwidth]{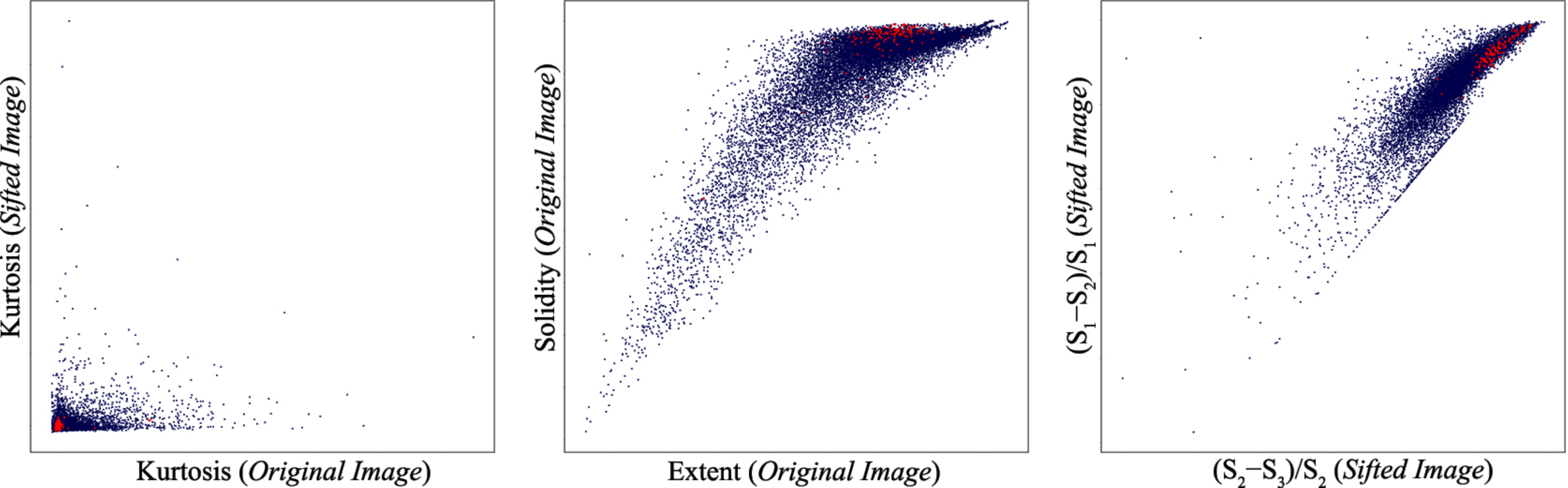}}
\caption{Primary features used to discriminate positive ($T$) and negative ($F$) candidates prior to feed to the classifier. $T$ is marked in red and $F$ is marked in blue.}
\end{figure}

\section{Discussion}
\subsection{Superpixel Contours}
In our model, SLIC extraction provides $99.1\%$ detection rate, failing to extract only a single mass. But it has one major setback with regards to the generated shape for all superpixel candidates. A high compact factor ensures better segmentation, but distorts the original contours of the candidates $-$an important feature for masses. Hence, although we achieve a significant Dice coefficient, the artificial contours likely cap feature extraction performance. In this regard, Chan Vese active contours can be used to refine the superpixel contours and preserve their boundary information \cite{11}.

\subsection{Role of Multi-Scale Sifting in Classification}
Morphological enhancement not only facilitated segmentation, but it also boosted the performance during classification. In fact, some of the most dominant discriminating features were derived from the sifted images, as shown in Fig. 4. These features were carefully administered to highlight all the basic characteristics of masses (shape, boundary contrast, texture and intensity).

\subsection{Flexibility of Cascaded Classifier}
Cascaded ensembles of SVMs played a key role in tackling the strong class imbalance. However, cascaded layers also allow ample flexibility to add specialized features in a later stage, at will. Calculating these heavier features on the entire population of superpixel candidates in an earlier layer or prior to classification, will otherwise be too computationally inefficient and time consuming. But they can prove useful to target the more difficult false positive candidates remaining in the final stages of the cascade. An example of such a complex feature can be "isocontours", as described by Byung-Woo Hong et al. \cite{10}, to create a topological model of breast masses.

\subsection{False Positive Prediction Analysis}
Before determining the optimum threshold, the final stage of the classifier produces a probability score for each candidate's likelihood to be a mass. By plotting these predictions in a density \textit{"heat map"} for each image, we can obtain a unique insight into the nature of our false positives. Fig. 5 demonstrates such an example. Here, the classifier is successful in predicting the correct candidate as a mass. But we can also see multiple potential false positives that can arise based on the threshold selected. Upon further inspection, it is clear where the classifier is struggling, as the closest false positive region in the original mammogram is normal dense tissue resembling a breast mass. However, since we use a multi-scale approach, each pixel in the original image receives a total of four prediction scores $-$one from each of the superpixels occupying the pixel in the sifted images. By combining these scores together, the classifier grasps a broad understanding of tissue regions and this may be the tipping point for the correct prediction.

\vspace{4mm}

\begin{figure}[h!]
\centerline{\includegraphics[width=0.48\textwidth]{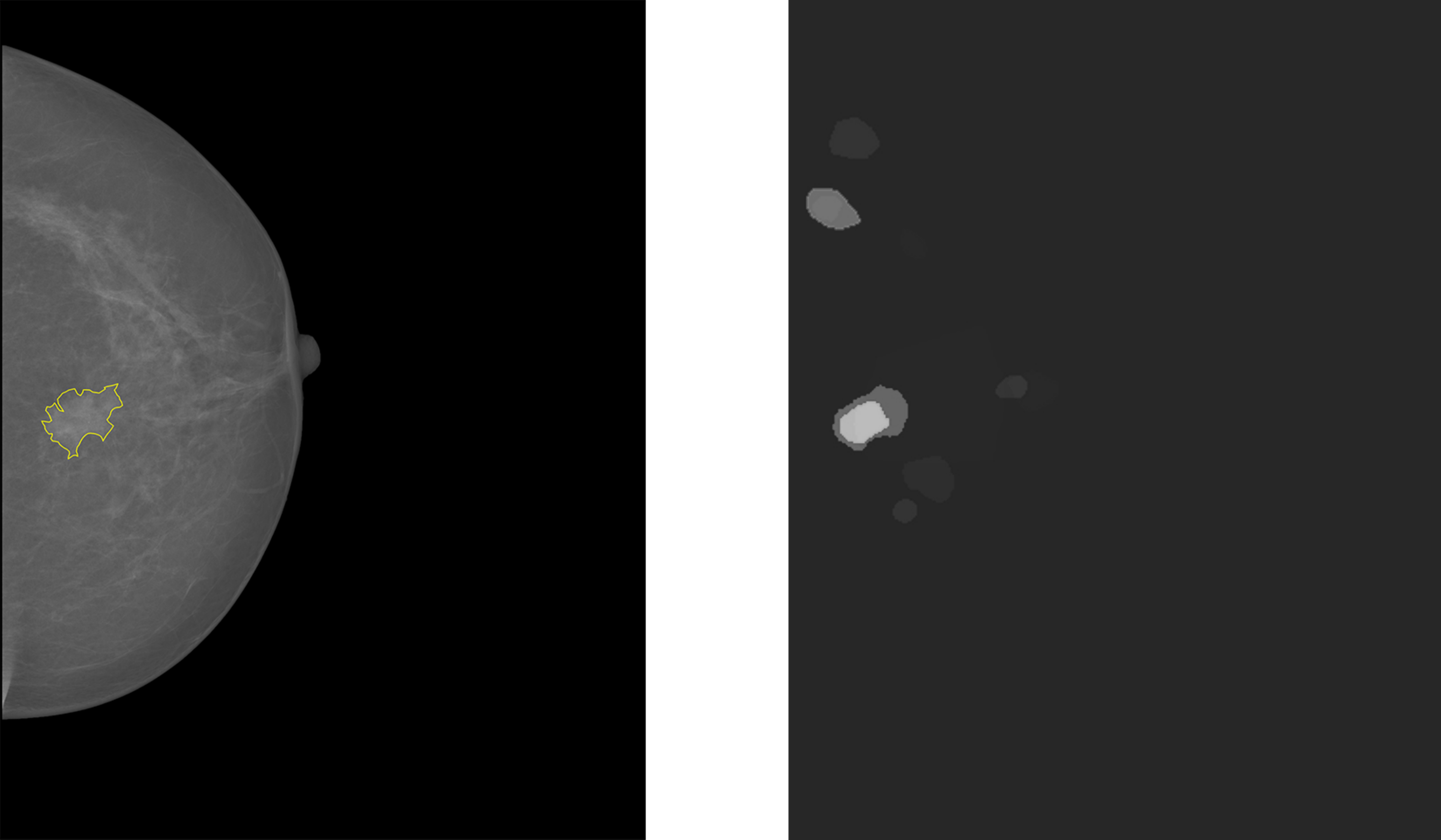}}
\caption{Screening mammogram example with its corresponding ground truth overlaid in yellow (left) and the prediction heat map derived from the final stage of the classifier (right).}
\end{figure}

\vspace{3mm}

\section{Conclusion}
An automated mass detection model for mammograms was proposed, trained and evaluated using the \textit{INbreast} dataset. The method was based on multi-scale morphological enhancement and SLIC superpixels algorithm to generate region candidates, and a supervised machine learning approach using three-stage cascaded ensemble of SVMs to classify candidates. After critical analysis, redesigning and calibrating across all primary parameters in the pipeline, a significant performance has been achieved using this architecture. Nonetheless, additional hand-crafted features can be incorporated and identified performance bottlenecks in the system can be addressed and improved upon in a future edition.

\vfill
\pagebreak

\end{document}